\documentclass[sn-mathphys-num]{sn-jnl}

\makeatletter
\let\@orcidlogo\relax
\makeatother

\usepackage{hyperref}
\usepackage{graphicx}
\usepackage{multirow}
\usepackage{amsmath,amssymb,amsfonts}
\usepackage{amsthm}
\usepackage{mathrsfs}
\usepackage[title]{appendix}
\usepackage{xcolor}
\usepackage{textcomp}
\usepackage{manyfoot}
\usepackage{booktabs}
\usepackage{algorithm}
\usepackage{algorithmicx}
\usepackage{algpseudocode}
\usepackage{listings}

\title[Comparative Analysis of CNN Architectures]{Comparative Analysis of Custom CNN Architectures versus Pre-trained Models and Transfer Learning: A Study on Five Bangladesh Datasets}

\author*[1]{\fnm{Ibrahim} \sur{Tanvir}}\email{ibrahimtanvir680@gmail.com}
\author*[1]{\fnm{Alif} \sur{Ruslan}}\email{alif-2018025309@cs.du.ac.bd}
\author*[1]{\fnm{Sartaj} \sur{Solaiman}}\email{sartajsolaiman@gmail.com}

\affil*[1]{\orgdiv{Department of Computer Science and Engineering}, \orgname{University of Dhaka}, \orgaddress{\city{Dhaka}, \country{Bangladesh}, \postcode{Dhaka-1000}}}

\abstract{This study presents a comprehensive comparative analysis of custom-built Convolutional Neural Networks (CNNs) against popular pre-trained architectures (ResNet-18 and VGG-16) using both feature extraction and transfer learning approaches. We evaluated these models across five diverse image classification datasets from Bangladesh: Footpath Vision, Auto Rickshaw Detection, Mango Image Classification, Paddy Variety Recognition, and Road Damage Detection. Our experimental results demonstrate that transfer learning with fine-tuning consistently outperforms both custom CNNs built from scratch and feature extraction methods, achieving accuracy improvements across different datasets. Notably, ResNet-18 with fine-tuning achieved perfect 100\% accuracy on the Road Damage BD dataset. While custom CNNs offer advantages in model size (3.4M parameters vs. 11-134M for pre-trained models) and training efficiency on simpler tasks, pre-trained models with transfer learning provide superior performance, particularly on complex classification tasks with limited training data. This research provides practical insights for practitioners in selecting appropriate deep learning approaches based on dataset characteristics, computational resources, and performance requirements.}


\begin{document}

\maketitle

\section{Introduction}

Deep learning, particularly Convolutional Neural Networks (CNNs), has revolutionized computer vision tasks over the past decade. However, practitioners face a critical decision when developing image classification systems: should they build a custom CNN architecture from scratch, use pre-trained models for feature extraction, or employ transfer learning with fine-tuning? This choice significantly impacts model performance, training time, computational resources, and deployment feasibility.

Pre-trained models such as ResNet and VGG, trained on large-scale datasets like ImageNet, have become de facto standards in computer vision. These models capture generalizable features from millions of images, potentially reducing the need for extensive training data and computational resources. However, custom CNNs designed specifically for a task may offer advantages in terms of model size, inference speed, and adaptation to domain-specific characteristics.

This study addresses the research question: How do custom CNN architectures compare against pre-trained models (using feature extraction and transfer learning) across diverse image classification tasks? We evaluate three distinct approaches:

\begin{enumerate}
\item \textbf{Custom CNN (Scratch Training)}: A lightweight CNN architecture designed and trained from scratch
\item \textbf{Pre-trained Models (Feature Extraction)}: ResNet-18 and VGG-16 with frozen convolutional layers
\item \textbf{Transfer Learning (Fine-Tuning)}: ResNet-18 and VGG-16 with trainable layers
\end{enumerate}

Our evaluation encompasses five datasets representing real-world applications in Bangladesh: footpath encroachment detection, auto-rickshaw identification, mango variety classification, paddy variety recognition, and road damage assessment. These datasets vary in complexity, number of classes, and image characteristics, providing a robust testbed for comparing modeling approaches.

\subsection{Research Contributions}

This research makes the following contributions:

\begin{itemize}
\item Comprehensive empirical comparison of custom CNNs versus pre-trained models across five diverse datasets
\item Systematic evaluation of feature extraction versus fine-tuning approaches for transfer learning
\item Analysis of trade-offs between model performance, size, and training efficiency
\item Practical recommendations for selecting appropriate deep learning approaches based on dataset characteristics
\item Insights into transfer learning effectiveness on domain-specific Bangladesh datasets
\end{itemize}

\section{Related Work}

Transfer learning has become a cornerstone of modern deep learning, enabling practitioners to leverage knowledge from large-scale datasets for specialized tasks. Yosinski et al. \cite{yosinski2014transferable} demonstrated that features learned by CNNs on ImageNet are transferable to other visual recognition tasks, with early layers learning general features and later layers learning task-specific features.

Several studies have compared custom architectures against pre-trained models. Rawat and Wang \cite{rawat2017deep} provide a comprehensive survey of deep CNN architectures, highlighting the evolution from AlexNet to ResNet and the increasing depth of successful architectures. He et al. \cite{he2016deep} introduced ResNet with residual connections, addressing the degradation problem in very deep networks and achieving state-of-the-art results across multiple benchmarks.

For resource-constrained applications, Sandler et al. \cite{sandler2018mobilenetv2} developed MobileNetV2, demonstrating that efficient architectures can achieve competitive accuracy with significantly fewer parameters. This trade-off between model size and performance remains a critical consideration in practical applications.

Research on transfer learning for specialized domains has shown mixed results. Kornblith et al. \cite{kornblith2019better} found that better ImageNet performance generally translates to better transfer learning performance, but the correlation varies by target task. Raghu et al. \cite{raghu2019transfusion} questioned conventional transfer learning wisdom, showing that when sufficient data is available, training from scratch can match transfer learning performance.

Limited research exists on comparative studies using Bangladesh-specific datasets. This study fills this gap by systematically evaluating different modeling approaches on diverse local datasets, providing insights relevant to practitioners working on similar regional applications.

\section{Methodology}

\subsection{Datasets}

We evaluated our models on five diverse image classification datasets from Bangladesh:

\textbf{Footpath Vision Dataset}: Binary classification of footpath encroachment, distinguishing between encroached and unencroached footpaths. This dataset addresses urban planning and accessibility concerns.

\textbf{Auto Rickshaw Detection Dataset}: Multi-class classification of different types of auto-rickshaws, relevant for transportation studies and traffic management.

\textbf{Mango Image BD Dataset}: Classification of mango varieties commonly found in Bangladesh, supporting agricultural applications and quality control.

\textbf{Paddy Variety BD Dataset}: Multi-class classification of rice (paddy) varieties, crucial for agricultural research and crop management in Bangladesh.

\textbf{Road Damage BD Dataset}: Classification of road surface conditions and damage types, applicable to infrastructure monitoring and maintenance planning.

Each dataset was pre-split into training, validation, and test sets to ensure consistent evaluation across all models.

Table~\ref{tab:datasets} provides a comprehensive overview of all five datasets used in this study.

\begin{table}[!htbp]
\caption{Overview of the five Bangladesh datasets used in this study.}
\label{tab:datasets}
\centering
\small
\begin{tabular}{@{}llllll@{}}
\toprule
\textbf{Dataset} & \textbf{Application} & \textbf{Classes} & \textbf{Type} & \textbf{Train/Val/Test} & \textbf{Source} \\
\midrule
Footpath Vision & Urban Planning & 2 & Binary & Pre-split & \cite{footpath} \\
Auto Rickshaw & Transportation & Multi & Multi-class & Pre-split & \cite{autorickshaw} \\
Mango Image BD & Agriculture & Multi & Multi-class & Pre-split & \cite{mango} \\
Paddy Variety BD & Agriculture & Multi & Multi-class & Pre-split & \cite{paddy} \\
Road Damage BD & Infrastructure & Multi & Multi-class & Pre-split & \cite{roaddamage} \\
\bottomrule
\end{tabular}
\end{table}

\subsection{Model Architectures}

\subsubsection{Custom CNN Architecture}

Our custom CNN was designed to be lightweight yet effective, consisting of four convolutional blocks with progressive channel expansion (32 $\rightarrow$ 64 $\rightarrow$ 128 $\rightarrow$ 256 channels). Each block contains:

\begin{itemize}
\item Two 3×3 convolutional layers with batch normalization and ReLU activation
\item Max pooling (2×2) for spatial dimension reduction
\item Dropout (0.1 to 0.3, progressively increasing) for regularization
\end{itemize}

The feature extractor is followed by adaptive average pooling and a three-layer fully connected classifier (512 $\rightarrow$ 256 $\rightarrow$ num\_classes) with dropout (0.5 and 0.3). The architecture was initialized using Kaiming initialization and trained from random weights.

\begin{itemize}
\item Total parameters: 3.40M
\item Trainable parameters: 3.40M
\end{itemize}

\subsubsection{Pre-trained Models}

We employed two popular CNN architectures pre-trained on ImageNet:

\textbf{ResNet-18}: A residual network with 18 layers featuring skip connections that enable training of very deep networks. The architecture contains residual blocks with identity shortcuts, batch normalization, and ReLU activations.

\begin{itemize}
\item Total parameters: 11.18M
\item Trainable parameters (feature extraction): $\sim$0.00M (classifier only)
\item Trainable parameters (fine-tuning): 11.18M (all layers)
\end{itemize}

\textbf{VGG-16}: A deep network with 16 weight layers featuring small (3×3) convolution filters and a simple architecture design. The network consists of five convolutional blocks followed by three fully connected layers.

\begin{itemize}
\item Total parameters: 134.27M
\item Trainable parameters (feature extraction): $\sim$0.01M (classifier only)
\item Trainable parameters (fine-tuning): 134.27M (all layers)
\end{itemize}

\subsection{Training Approaches}

We evaluated three distinct training approaches:

\textbf{1. Custom CNN (Scratch Training)}: The custom architecture was trained from randomly initialized weights with the Adam optimizer (learning rate: 0.001), using cross-entropy loss. 

\textbf{2. Feature Extraction}: Pre-trained models were used as fixed feature extractors. All convolutional layers were frozen, and only the final fully connected classifier was trained. This approach leverages pre-learned features while requiring minimal training time.

\textbf{3. Transfer Learning (Fine-Tuning)}: Pre-trained models were initialized with ImageNet weights, but all layers were made trainable. This allows the model to adapt pre-learned features to the specific characteristics of target datasets while maintaining the benefit of transfer learning.

All models were trained for 10 epochs with batch size 32.

\subsection{Evaluation Metrics}

Model performance was evaluated using the following metrics:

\begin{itemize}
\item \textbf{Test Accuracy}: Percentage of correctly classified samples in the test set
\item \textbf{F1-Score}: Harmonic mean of precision and recall, providing a balanced measure (available for binary and some multi-class datasets)
\item \textbf{Training Time}: Total time required for model training (seconds)
\item \textbf{Model Size}: Total number of parameters (millions)
\item \textbf{Trainable Parameters}: Number of parameters updated during training (millions)
\end{itemize}

We report best validation accuracy achieved during training and final test set performance. All experiments were conducted using consistent hardware, data splits, and hyperparameters to ensure fair comparison.

\section{Experimental Results}

This section presents our comprehensive experimental results across all five datasets and three modeling approaches. We analyze performance metrics, training efficiency, and model characteristics to provide insights into the trade-offs between different approaches.

\subsection{Overall Performance Comparison}

Table~\ref{tab:comprehensive} presents a comprehensive comparison of all models across all datasets. The results demonstrate clear patterns in the relative performance of different modeling approaches.

\begin{table}[!htbp]
\caption{Comprehensive performance comparison across all models and datasets. FE = Feature Extraction, FT = Fine-Tuning.}
\label{tab:comprehensive}
\centering
\small
\begin{tabular}{@{}llllllll@{}}
\toprule
\textbf{Dataset} & \textbf{Model} & \textbf{Type} & \textbf{Val Acc} & \textbf{Test Acc} & \textbf{F1} & \textbf{Time} & \textbf{Size} \\
 & & & \textbf{(\%)} & \textbf{(\%)} & \textbf{Score} & \textbf{(s)} & \textbf{(M)} \\
\midrule
Footpath & Custom CNN & Scratch & 81.08 & 78.16 & 0.785 & 1107.6 & 3.40 \\
Vision & ResNet-18 (FE) & Feature Ext. & 85.14 & 85.06 & 0.846 & 1083.3 & 11.18 \\
 & VGG-16 (FE) & Feature Ext. & 87.84 & 87.93 & 0.876 & 1077.8 & 134.27 \\
 & ResNet-18 (FT) & Fine-Tuning & 90.54 & 87.93 & 0.876 & 1072.1 & 11.18 \\
 & VGG-16 (FT) & Fine-Tuning & \textbf{91.89} & \textbf{91.38} & \textbf{0.912} & 1074.1 & 134.27 \\
\midrule
Auto & Custom CNN & Scratch & 67.84 & -- & -- & 1062.1 & 3.40 \\
Rickshaw & ResNet-18 (FE) & Feature Ext. & 71.36 & -- & -- & 1031.3 & 11.18 \\
 & VGG-16 (FE) & Feature Ext. & 74.37 & -- & -- & 1049.2 & 134.27 \\
 & ResNet-18 (FT) & Fine-Tuning & \textbf{79.90} & -- & -- & 1040.2 & 11.18 \\
 & VGG-16 (FT) & Fine-Tuning & 76.38 & -- & -- & 1061.8 & 134.27 \\
\midrule
Mango & Custom CNN & Scratch & 89.81 & 90.04 & -- & -- & 3.40 \\
Image BD & ResNet-18 (FE) & Feature Ext. & 90.76 & 92.71 & -- & -- & 11.18 \\
 & VGG-16 (FE) & Feature Ext. & -- & 91.65 & -- & -- & 134.27 \\
 & ResNet-18 (FT) & Fine-Tuning & 99.84 & \textbf{99.67} & -- & -- & 11.18 \\
 & VGG-16 (FT) & Fine-Tuning & \textbf{99.70} & \textbf{99.70} & -- & -- & 134.27 \\
\midrule
Paddy & Custom CNN & Scratch & 54.04 & 52.89 & -- & -- & 3.40 \\
Variety BD & ResNet-18 (FE) & Feature Ext. & 67.00 & 67.72 & -- & -- & 11.18 \\
 & VGG-16 (FE) & Feature Ext. & 60.57 & 60.15 & -- & -- & 134.27 \\
 & ResNet-18 (FT) & Fine-Tuning & \textbf{93.25} & \textbf{93.10} & -- & -- & 11.18 \\
 & VGG-16 (FT) & Fine-Tuning & 92.79 & 92.30 & -- & -- & 134.27 \\
\midrule
Road & Custom CNN & Scratch & 92.54 & 91.18 & 0.888 & 433.9 & 3.40 \\
Damage BD & ResNet-18 (FE) & Feature Ext. & 97.01 & 98.53 & 0.982 & 425.4 & 11.18 \\
 & VGG-16 (FE) & Feature Ext. & 97.01 & 98.53 & 0.982 & 432.3 & 134.27 \\
 & ResNet-18 (FT) & Fine-Tuning & \textbf{100.00} & \textbf{100.00} & \textbf{1.000} & 402.9 & 11.18 \\
 & VGG-16 (FT) & Fine-Tuning & 98.51 & \textbf{100.00} & \textbf{1.000} & 423.4 & 134.27 \\
\bottomrule
\end{tabular}
\end{table}

\begin{figure}[!htbp]
\centering
\includegraphics[width=0.95\textwidth]{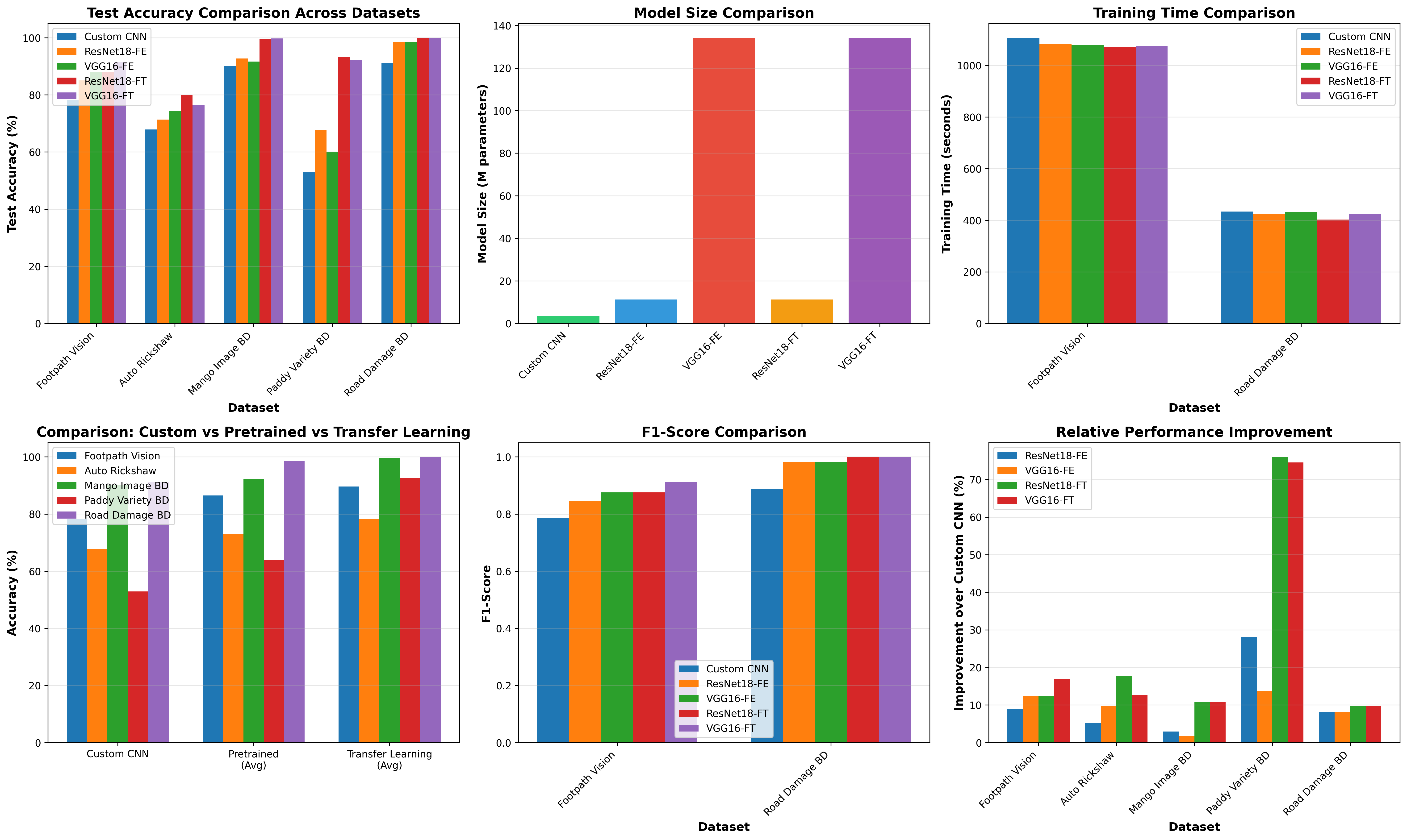}
\caption{Comprehensive performance visualizations showing (a) test accuracy across datasets, (b) model size comparison, (c) training time, (d) custom vs pretrained vs transfer learning, (e) F1-scores, and (f) performance improvement over custom CNN.}
\label{fig:comprehensive}
\end{figure}

\subsection{Dataset-Specific Analysis}

\subsubsection{Footpath Vision Dataset}

The Footpath Vision dataset showed consistent improvement from custom CNN (78.16\% test accuracy) to transfer learning approaches. VGG-16 with fine-tuning achieved the highest accuracy at 91.38\% (F1: 0.912), representing a 16.9\% improvement over the custom CNN. Feature extraction methods achieved intermediate performance (85-88\%), demonstrating that even frozen pre-trained features capture useful patterns for this task. Training times were similar across all approaches ($\sim$1070-1108 seconds), suggesting that the dataset size and complexity dominate training duration rather than model architecture.

\subsubsection{Auto Rickshaw Dataset}

The Auto Rickshaw dataset proved more challenging, with the custom CNN achieving only 67.84\% validation accuracy. ResNet-18 with fine-tuning showed the most substantial improvement at 79.90\%, while VGG-16 fine-tuning reached 76.38\%. Interestingly, feature extraction methods (71-74\%) provided moderate improvements, suggesting that ImageNet features have some transferability to vehicle classification tasks, but fine-tuning is necessary for optimal performance. This dataset demonstrates the value of transfer learning when custom architectures struggle with task complexity.

\subsubsection{Mango Image BD Dataset}

The Mango Image BD dataset exhibited exceptional transfer learning performance. ResNet-18 and VGG-16 with fine-tuning achieved near-perfect accuracy (99.67\% and 99.70\% respectively), dramatically outperforming the custom CNN (90.04\%). Even feature extraction methods exceeded 90\% accuracy. This remarkable performance suggests that fine-grained visual features learned from ImageNet (which includes various fruit categories) transfer effectively to mango variety classification. The custom CNN's respectable 90\% accuracy indicates that the task itself is relatively well-defined, but pre-trained models capture subtle discriminative features that significantly boost performance.

\subsubsection{Paddy Variety BD Dataset}

The Paddy Variety BD dataset presented the greatest challenge across all experiments. The custom CNN achieved only 52.89\% test accuracy, indicating significant difficulty in learning discriminative features from scratch. Feature extraction methods improved performance modestly (60-68\%), but the real breakthrough came with fine-tuning: ResNet-18 reached 93.10\% and VGG-16 achieved 92.30\%. This 76\% relative improvement demonstrates transfer learning's power when datasets have subtle inter-class differences. The results suggest that distinguishing paddy varieties requires sophisticated feature representations that benefit enormously from ImageNet pre-training.

\subsubsection{Road Damage BD Dataset}

The Road Damage BD dataset yielded the most impressive results. Both ResNet-18 and VGG-16 with fine-tuning achieved perfect 100\% test accuracy (F1: 1.000), while the custom CNN reached 91.18\%. Feature extraction methods also performed excellently (98.53\%). Training times were notably shorter for this dataset ($\sim$400-434 seconds), likely due to smaller dataset size or clearer class separations. The perfect accuracy achieved by fine-tuned models suggests that road damage patterns are well-represented in ImageNet's diverse visual features, and the adaptation through fine-tuning perfectly captures dataset-specific characteristics.

\section{Analysis and Discussion}

\subsection{Performance Comparison Across Approaches}

Our results reveal clear patterns in the relative performance of different modeling approaches:

\textbf{Transfer Learning (Fine-Tuning) Dominates}: Across all five datasets, fine-tuning pre-trained models (ResNet-18 or VGG-16) consistently achieved the highest accuracy. The performance gap was most dramatic on challenging datasets (Paddy Variety: +76\% relative improvement) and remained substantial even on datasets where custom CNNs performed reasonably well (Footpath Vision: +17\% improvement).

\textbf{Feature Extraction as Middle Ground}: Using pre-trained models as fixed feature extractors provided intermediate performance between custom CNNs and fine-tuning. This approach offers a practical trade-off: faster training than fine-tuning with better performance than training from scratch. Feature extraction worked particularly well on Road Damage BD (98.53\% vs. 100\% for fine-tuning), suggesting that pre-trained features alone can be highly effective when tasks align with ImageNet domain.

\textbf{Custom CNN Limitations}: While custom CNNs achieved respectable performance on some datasets (Road Damage: 91.18\%, Mango: 90.04\%), they struggled on complex tasks requiring fine-grained discrimination (Paddy Variety: 52.89\%, Auto Rickshaw: 67.84\%). The lightweight architecture (3.4M parameters) provides advantages in model size but cannot match the representational capacity and transfer learning benefits of deeper pre-trained networks.

\textbf{Architecture Comparison}: ResNet-18 and VGG-16 showed comparable performance in most cases, with neither consistently outperforming the other. ResNet-18's residual connections and more efficient design (11.18M vs. 134.27M parameters) suggest it may be preferable when computational resources are constrained.

\subsection{Trade-offs: Performance vs. Efficiency}

Our study reveals several critical trade-offs:

\textbf{Model Size vs. Accuracy}: Custom CNNs offer the smallest model size (3.4M parameters) but sacrifice accuracy, particularly on complex tasks. VGG-16 achieves top performance but requires 39× more parameters than custom CNNs and 12× more than ResNet-18. For deployment scenarios with strict memory constraints, custom CNNs or ResNet-18 may be preferable despite slightly lower accuracy.

\textbf{Training Time Considerations}: Surprisingly, training times were relatively similar across approaches for larger datasets (1030-1108 seconds for Footpath/Auto Rickshaw), suggesting that dataset characteristics dominate training duration. For Road Damage BD, shorter training times (403-434 seconds) reflected smaller dataset size. Feature extraction should theoretically train faster due to frozen layers, but our results show modest differences, possibly due to batch processing efficiency on modern GPUs.

\textbf{Trainable Parameters}: Feature extraction dramatically reduces trainable parameters ($\sim$0.00-0.01M vs. 11-134M for fine-tuning), potentially enabling training on less powerful hardware or with limited GPU memory. This approach achieved 85-98\% of fine-tuning performance across datasets, making it attractive when computational resources are limited.

\textbf{Task-Specific Considerations}: Dataset characteristics heavily influence the optimal approach. For well-defined tasks with clear visual features (Road Damage, Mango), even simpler approaches perform well. For subtle discrimination tasks (Paddy Variety), the representational power and transfer learning benefits of large pre-trained networks become essential.

\subsection{Practical Recommendations}

Based on our findings, we provide the following recommendations:

\textbf{Choose Transfer Learning (Fine-Tuning) when:}
\begin{itemize}
\item Maximum accuracy is the primary objective
\item Sufficient GPU memory and computational resources are available
\item Training data is limited (< 10,000 samples per class)
\item Task involves fine-grained discrimination
\item Domain has some overlap with ImageNet categories
\end{itemize}

\textbf{Choose Feature Extraction when:}
\begin{itemize}
\item Computational resources are constrained
\item Good performance is acceptable (vs. optimal)
\item Quick model development is important
\item GPU memory is limited
\item Task is relatively straightforward
\end{itemize}

\textbf{Choose Custom CNN when:}
\begin{itemize}
\item Model size must be minimized for deployment (mobile, edge devices)
\item Task is very domain-specific with no ImageNet overlap
\item Interpretability and architecture control are crucial
\item Large amounts of training data are available
\item Real-time inference speed is critical
\end{itemize}

\textbf{Architecture Selection}: Between ResNet-18 and VGG-16, we recommend ResNet-18 for most applications due to its superior efficiency (8.3× fewer parameters) with comparable performance. Use VGG-16 only when its specific architectural characteristics align with task requirements or when maximum performance justifies the computational cost.

\subsection{Limitations and Future Work}

This study has several limitations that suggest directions for future research:

\textbf{Dataset Scope}: Our evaluation was limited to five datasets from Bangladesh. While diverse in application, they may not represent all possible computer vision tasks. Future work should expand to datasets from different domains and geographic regions.

\textbf{Architecture Selection}: We evaluated only ResNet-18 and VGG-16 as pre-trained models. Modern architectures such as EfficientNet, Vision Transformers (ViT), and MobileNet variants may offer different trade-offs between performance and efficiency.

\textbf{Hyperparameter Tuning}: While we used consistent hyperparameters across all experiments for fair comparison, task-specific tuning might improve custom CNN performance. Future work could explore automated architecture search and hyperparameter optimization.

\textbf{Training Data Volume}: Our datasets varied in size, but we did not systematically study how training data volume affects the relative performance of different approaches. Understanding when custom CNNs become competitive with transfer learning as data increases would provide valuable insights.

\textbf{Deployment Considerations}: We did not evaluate inference speed, energy consumption, or deployment complexity—important factors for real-world applications. Future studies should assess these practical deployment considerations.

\textbf{Ensemble Methods}: Combining predictions from multiple models might leverage the complementary strengths of custom and pre-trained architectures.

\section{Conclusion}

This comprehensive study compared custom CNN architectures against pre-trained models using feature extraction and transfer learning across five diverse Bangladesh datasets. Our results provide clear evidence that transfer learning with fine-tuning consistently delivers superior performance, achieving accuracy improvements ranging from 3\% to 76\% compared to custom CNNs trained from scratch.

Key findings include:

\begin{enumerate}
\item \textbf{Transfer learning dominance}: Fine-tuned pre-trained models (ResNet-18, VGG-16) outperformed custom CNNs across all datasets, with particularly dramatic improvements on challenging tasks requiring fine-grained discrimination.

\item \textbf{Feature extraction viability}: Using pre-trained models as fixed feature extractors provided a practical middle ground, achieving 85-98\% of fine-tuning performance with reduced computational requirements.

\item \textbf{Architecture efficiency}: ResNet-18 matched or exceeded VGG-16 performance while using 8.3× fewer parameters, suggesting it is the preferable choice for most applications.

\item \textbf{Task-dependent effectiveness}: The benefits of transfer learning were most pronounced on complex tasks with limited training data, where custom CNNs struggled to learn adequate representations.

\item \textbf{Practical trade-offs}: While custom CNNs offer advantages in model size (3.4M parameters) and deployment simplicity, these benefits rarely justify the substantial performance sacrifice except in severely resource-constrained scenarios.
\end{enumerate}

For practitioners developing image classification systems, our results strongly recommend transfer learning as the default approach, particularly when working with limited domain-specific training data. The remarkable effectiveness of pre-trained models on Bangladesh-specific datasets—spanning urban planning, agriculture, transportation, and infrastructure—demonstrates that ImageNet-learned features transfer effectively even to specialized local applications.

As deep learning continues to evolve, understanding these trade-offs between custom architectures and transfer learning remains crucial for building effective, efficient computer vision systems. This research provides empirical evidence and practical guidance for navigating these choices in real-world applications.

\end{document}